\pdfoutput=1

\def\year{2020}\relax
\documentclass[letterpaper]{article} % DO NOT CHANGE THIS
\usepackage{aaai20}  % DO NOT CHANGE THIS
\usepackage{times}  % DO NOT CHANGE THIS
\usepackage{helvet} % DO NOT CHANGE THIS
\usepackage{courier}  % DO NOT CHANGE THIS
\usepackage[hyphens]{url}  % DO NOT CHANGE THIS
\usepackage{graphicx} % DO NOT CHANGE THIS
\usepackage{graphicx}  % DO NOT CHANGE THIS
\usepackage[utf8]{inputenc} % allow utf-8 input
\usepackage[T1]{fontenc}    % use 8-bit T1 fonts
\usepackage{url}            % simple URL typesetting
\usepackage{booktabs}       % professional-quality tables
\usepackage{amsfonts}       % blackboard math symbols
\usepackage{nicefrac}       % compact symbols for 1/2, etc.
\usepackage{microtype}      % microtypograph
\usepackage{amsmath}
\usepackage{amssymb}
\usepackage{multirow}
\usepackage[ruled]{algorithm2e}
\usepackage{acronym}
\usepackage{enumitem}
\usepackage{subcaption}
\usepackage[dvipsnames]{xcolor}
\usepackage{graphicx}
\usepackage{verbatim}

\newcommand{\citet}[1]{\citeauthor{#1} \citeyear{#1}}

\makeatletter
\DeclareRobustCommand\onedot{\futurelet\@let@token\@onedot}
\def\@onedot{\ifx\@let@token.\else.\null\fi\xspace}
\def\eg{\emph{e.g}\onedot}

\makeatother

\title{Machine Number Sense:\\A Dataset of Visual Arithmetic Problems for Abstract and Relational Reasoning}
\author{
    Wenhe Zhang,\textsuperscript{\rm 1,3}
    Chi Zhang,\textsuperscript{\rm 1,2}
    Yixin Zhu,\textsuperscript{\rm 1,2}
    Song-Chun Zhu\textsuperscript{\rm 1,2} \\
    \textsuperscript{\rm 1}UCLA Center for Vision, Cognition, Learning, and Autonomy \\
    \textsuperscript{\rm 2}International Center for AI and Robot Autonomy (CARA) \\
    \textsuperscript{\rm 3}Peking University \\
    wenhe@pku.edu.cn, chi.zhang@ucla.edu, yixin.zhu@ucla.edu, sczhu@stat.ucla.edu
}

\acrodef{aog}[AOG]{And-Or Graph}
\acrodef{mns}[MNS]{Machine Number Sense}

\begin{document}
\maketitle

\begin{abstract}
As a comprehensive indicator of mathematical thinking and intelligence, the \emph{number sense} \cite{dehaene2011number} bridges the induction of symbolic concepts and the competence of problem-solving. To endow such a crucial cognitive ability to machine intelligence, we propose a dataset, \acf{mns}, consisting of \emph{visual} arithmetic problems automatically generated using a grammar model---\acf{aog}. These visual arithmetic problems are in the form of geometric figures: each problem has a set of geometric shapes as its context and embedded number symbols. Solving such problems is not trivial; the machine not only has to recognize the number, but also to interpret the number with its contexts, shapes, and relations (\eg, symmetry) together with proper operations. We benchmark the \ac{mns} dataset using four predominant neural network models as baselines in this visual reasoning task. Comprehensive experiments show that current neural-network-based models still struggle to understand number concepts and relational operations. We show that a simple brute-force search algorithm could work out some of the problems without context information. Crucially, taking geometric context into account by an additional perception module would provide a sharp performance gain with fewer search steps. Altogether, we call for attention in fusing the classic search-based algorithms with modern neural networks to discover the essential number concepts in future research.
\end{abstract}

\section{Introduction}

\begin{quote}
    Number is the ruler of forms and ideas, and the cause of gods and demons.

    \hfill --- Pythagoras, c. 300 \cite{taylor1818nicomachean}
\end{quote}

Mathematics is arguably the most elegant and vivid reflection of human intelligence, covering the areas of geometry, arithmetic, algebra, and analysis \cite{dictionary1989oxford}. It is the science of logic reasoning, the discipline of abstract forms, and the realm of symbolic languages. Among all the mathematical symbols, numbers are the most familiar and vital elements to us. Although the opinions of Pythagoras that ``all is number'' are controversial and extreme, the significance of the numbers can never be overestimated: people from all walks of life embrace numbers every day.

Dealing with numbers seems to be a simple task and an innate competence: even newborn infants can discriminate basic numerosities, expressing their surprise when the number of stimuli changes from two to three \cite{starkey1980perception}. Meanwhile, processing numbers is also a painstaking challenge and a learned skill; it always takes years of efforts for students to practice calculations and more complicated computations. Such a numerical competence is, in fact, very unique; only few other animals possess similar capabilities (and at a much smaller scale compared to human) \cite{davis1988numerical,gallistel1989animal,gallistel1990organization,brannon1998ordering,dehaene1998abstract,cantlon2007much,jacob2008abc,nieder2009representation}. What is the underlying mechanism of human numerical thinking and the concepts of number? And how to endow a similar capability to machine intelligence?

The \emph{number sense} \cite{dehaene2011number}, a psychological terminology, provides an explanation about the cognitive process of numbers for both human and animals. It refers to the understanding of number concepts, the competence of numerical operations (including counting, comparison, estimation, and calculation), and the ability to flexibly solve mathematical problems \cite{bobis1996visualisation}. People characteristic of good number sense usually possess the abilities of fluent magnitude perception, reasonable result expectation, flexible mental computation, and appropriate presentation formulation \cite{kalchman2001psychological}. Below, we summarize four key observations from the vast body of literature on number sense.

\paragraph{Learned \emph{vs.} Innate}

Number sense is developed in \emph{acquired} environments in addition to our \emph{innate} capability. Five-month-old infants have already possessed the capacity to represent cardinality and can engage in rudimentary arithmetics---basic addition and subtraction operations on small sets of objects \cite{wynn1992addition}. Older children gradually \emph{learn} to establish the abstract connections between the magnitude of the quantities and the symbolic expression of the numbers, which are the foundation of further comparisons and calculations \cite{temple1998brain}. Symbolic numerical processing skills, different from the processing abilities of countable non-symbolic objects, are more closely related to mathematical competence \cite{schneider2017associations}. As for average adults, retrieving the abstract meaning of number symbols has been developed into a highly automated process, thus facilitating more rigorous computations \cite{dehaene2011number}.

\paragraph{Vision \emph{vs.} Language}

Number sense is in closer relation to vision than language due to a few reasons. First, the definition of number sense emphasizes the estimation of  the magnitude of the quantities and the understanding of number symbols based on \emph{visual} input. Second, empirical evidence has suggested the significant relation between vision and number sense. Studies of developmental psychology indicated that people first developed their number sense from vision; babies with limited knowledge of language have expressed the ability to discriminate the numerosity of \emph{visual} objects, and children gradually learn the quantitative meaning of \emph{visual} symbols \cite{dehaene2011number}. Third, in evolutionary psychology, animals, in general, are unable to generate a verbal representation of numbers, but some of them still exhibit the number sense, capable of numerical discrimination and mental operations of \emph{visual} items \cite{Gallistel2003Animal}.

\paragraph{Context and Adaptation}

Number sense is not only the awareness and manipulation of abstract symbols but also the capacity of conducting flexible mathematical operations in concrete situations. People with good number sense usually display an excellent problem-solving ability \cite{cobb1991assessment}. To solve mathematical problems effectively, we need to observe the \emph{context} in which the problem is presented, form an \emph{adaptive} representation for problem settings and a proper expectation for possible results, select the most suitable strategy that contains necessary sub-operations, and work with the numbers step by step \cite{heinze2009flexible}.

\paragraph{Quantity \emph{vs.} Rank}

Two types of neuronal mechanisms were extensively studied in the neuroscience literature \cite{wiese2003numbers}: (i) \emph{Numerical quantity} refers to the property of cardinality of sets of objects or events (also called numerosity)---``how many?''. (ii) \emph{Numerical rank} refers to the property of serial order and pertains to the question---``which position?''.

\subsection{Overview}

In this paper, we hope to use the concept of number sense, an ideal indicator, to evaluate the machine intelligence from the perspective of mathematics; it naturally combines both crystallized intelligence (knowledge and experience of number processing) and fluid intelligence (adaptive problem-solving in a given situation), which comprises the basic structure of human intelligence \cite{cattell1963theory}.

\begin{figure}[t!]
    \centering
    \includegraphics[width=\linewidth]{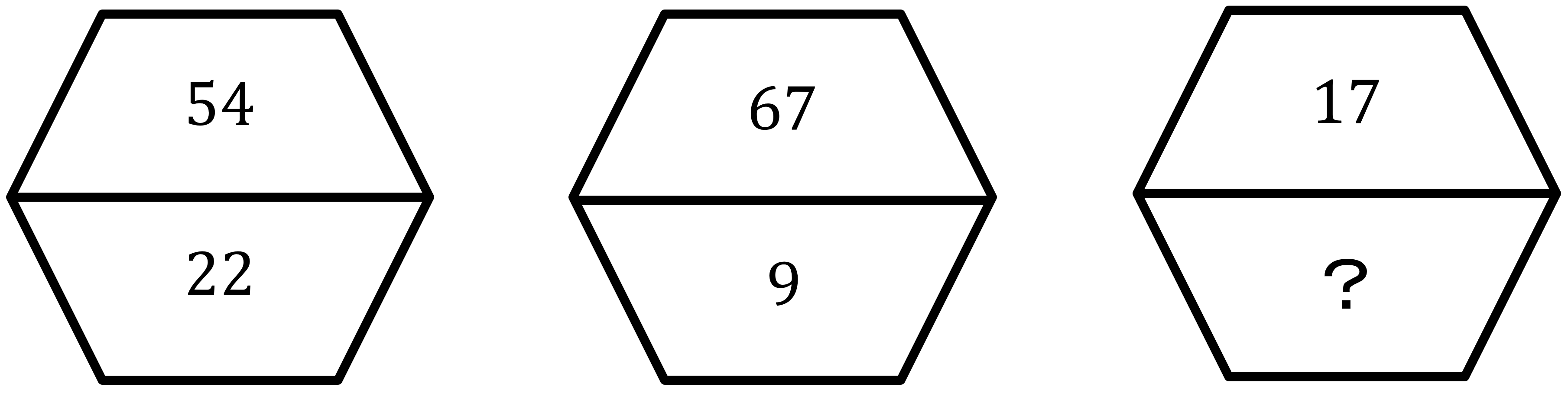}
    \caption{A sample problem in the \acf{mns} dataset using the rule of addition: $54 + 22 = 76$, $67 + 9 = 76$, $17 + ? = 76$. The correct answer is $59$.}
    \label{fig:example}
\end{figure}

Specifically, we propose a new dataset, \acf{mns}, in the form of geometric figures. It consists of various types of arithmetic problems, in which integers appear as problem contents and geometric shapes serve as problem contexts; see an example in Figure~\ref{fig:example}. The task for evaluating machine's number sense is: given training samples as \emph{images}, the algorithms should figure out the underlying latent relations between the numbers in each image panel and fill in the missing number as the answer in the last panel.

Here, we are interested in testing a few intriguing questions that correspond to the above four key observations: (i) Given only the visual input, are modern machine learning methods capable of learning and understanding the quantitative meaning of number symbols and the relations between these symbols? (ii) If the spaces of the operations and rules are known, is it possible to work out the problem by symbolic search? What would be the difference between the two streams of methods? (iii) How much does the contextual information contribute to numerical problem-solving? (iv) Could learning-based methods realize the numerical quantity and numerical rank merely from the visual input?

Our experiments show that the predominant neural network models still have a significant cognitive gap between visual symbols and abstract meanings even after extensive training; there must be a missing association between context information and problem-solving skill. In contrast, only taking number symbols as the input, the classic search algorithm manages to solve some problems correctly, but the search is very inefficient. Adding an additional perception module to provide geometric contextual information significantly improves the performance of the algorithm.

\begin{figure*}[t!]
    \centering
    \includegraphics[width=\linewidth]{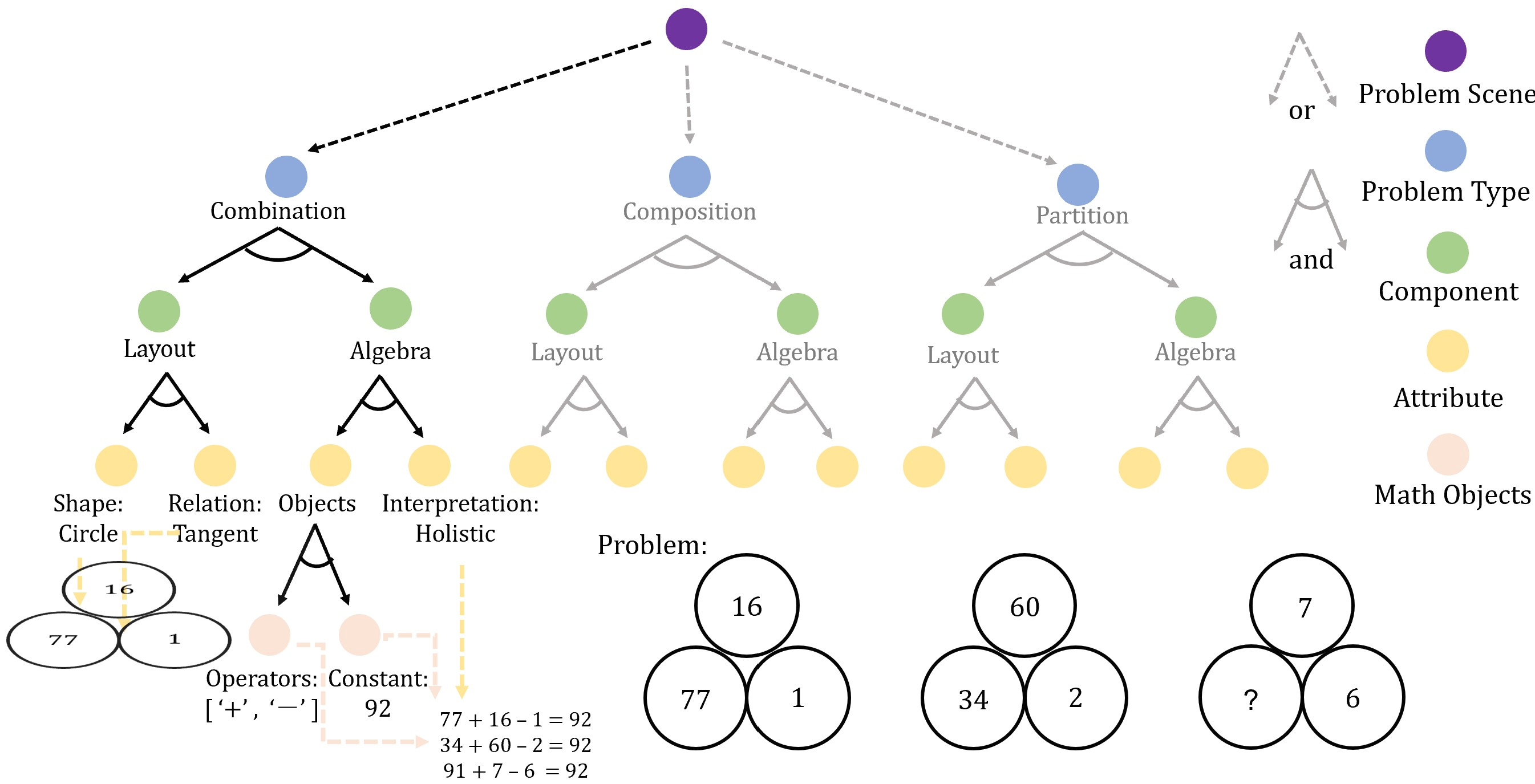}
    \caption{The \acf{mns} dataset creation process. Given grammar production rules together with its attributes, we can generate a test by parsing and sampling an \acf{aog}.}
    \label{fig:and_or_tree}
\end{figure*}

\subsection{Contributions}

This paper makes two major contributions:
\begin{itemize}[leftmargin=*,noitemsep,nolistsep]
    \item We introduce a new \acf{mns} dataset, composed of various visual arithmetic problems.
    \item We benchmark the ability of the modern machine learning methods with respect to the quantitative understanding of number symbols, the relational operations between numbers, and the ability of adaptive problem-solving, which together construct a full framework of number sense.
\end{itemize}

Compared to other mathematical problems in the form of text or language in prior work, the problems presented here are unique in the following aspects:

\paragraph{Token \emph{vs.} Pixel}

Instead of using tokenized symbols extracted from the texts or languages, testing machine number sense directly from pixel input is much more challenging. By means of language, the quantitative meaning of number symbols and the relations among them could be easily discovered with abundant semantic clues embedded in the sentence. In contrast, it is much harder to establish the connections among different numbers with their visual contexts; the algorithm has to reason and induce from the observed visual pattern using limited examples.

\paragraph{Sequential \emph{vs.} Hierarchical}

Using visual inputs also brings up more rigorous requirements for formulating suitable yet flexible representations. The visual pattern is usually hierarchically organized and generated, demanding an algorithm to parse a test into a similar hierarchical representation. This unique property is fundamentally different from the sequential and temporal nature in prior representations in the context of texts or languages. A proper perceptual-grouping (\eg, Gestalt laws \cite{wertheimer1923laws}) for visual elements is necessary. Additionally, we would need a flexible representation for representing a problem based on its context and reconstructing the representation when it is not appropriate; such an adaptation is regarded as a key step for problem-solving \cite{knoblich1999constraint}.

\paragraph{Recognition \emph{vs.} Reasoning}

The proposed dataset is characterized by both its simplicity and difficulty. In each problem, there are only numbers and geometric shapes, unlike others with various mathematical symbols \cite{ling2017program,saxton2019analysing}. However, simple appearance does not indicate trivial problem-solving; in contrast, it enforces the algorithm to reason about the latent structure, relations, and operations within a problem consisting of very ``limited'' visual information, making the problem-solving process challenging. This nature of the present dataset leads to the focus on reasoning and understanding, rather than the traditional tasks (\eg, recognition) in the field of computer vision.

\paragraph{Human \emph{vs.} Machine}

There are qualitative differences between the present dataset and previous tests of human number sense. The human tests examine number sense from a clinical perspective, aiming at discriminating children with potential mathematical disabilities, so that the problems in the tests are relatively easy, basic, and eliminative, serving as diagnostic tools. In contrast, our task investigates number sense from a cognitive perspective, measuring machine intelligence from the aspect of number processing; the problems are more comprehensive, flexible, and cognitive-demanding.

\section{The Machine Number Sense Dataset}

\paragraph{Representation}

We use \acf{aog} as the representation; see an illustration of the structure for the generation process in Figure~\ref{fig:and_or_tree}. \ac{aog} is a context-free grammar frequently used for hierarchical and compositional data in AI and computer vision \cite{zhu2007stochastic}. In \ac{mns} dataset, each problem has an internal hierarchical tree structure composed of And-nodes and Or-nodes; an And-node denotes a decomposition of a larger entity in the grammar, and an Or-node denotes an alternative decomposition. 

In our design, the root node of the \ac{aog} is an Or-node, representing a single test. After the decomposition on the sub-level are three different problem types represented by And-nodes. After selecting the problem type by choosing one of the And-nodes, the problem is divided into layout and algebra components. Sampling the terminal nodes in each component will complete the process of the problem generation. The three image panels within a single problem share common layout and algebraic properties; the only difference among them is the actual integers that appear on the panel.

\paragraph{Problem Types}

\begin{figure}[t!]
    \centering
    \includegraphics[width=\linewidth]{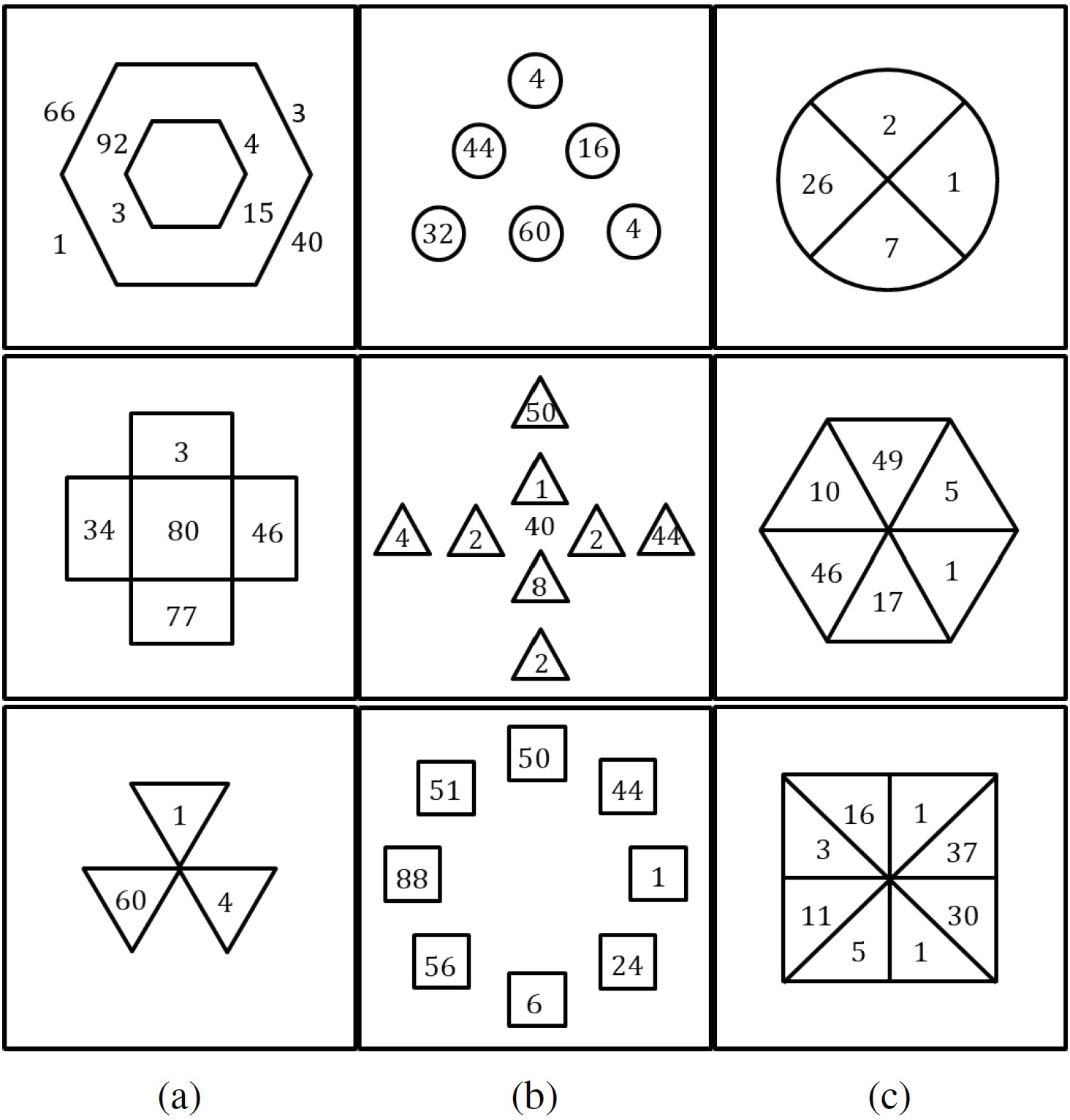}
    \caption{Layouts of three different problem types: (a) combination,  (b) composition, and (c) partition.}
    \label{fig:problem_examples}
\end{figure}

We design three types of problems: combination, composition, and partition, each of which has a distinctive layout. Figure~\ref{fig:problem_examples} shows a few examples using different layouts. In a combination problem, two or three geometric shapes are combined together by a specific spatial relation. In a composition problem, a set of small geometric shapes are composited to outline a larger shape. In a partition problem, one geometric shape is divided into several parts by lines.

\paragraph{Layout Component and Attributes}

The layout component serves as the problem \emph{context}, consisting of two different geometric attributes, both of which are necessary for the problem generation; see an illustration in Figure~\ref{fig:problem_examples}. The first attribute refers to geometric shape: triangle, square, circle, hexagon, or rectangle. The second attribute varies in different problem types. In combination problems, it indicates the spatial relation by which the geometric shapes group together; in our dataset, two figures could be combined by the relation of overlapping or including, and three figures could be grouped together by tangent relation. In composition problems, the second attribute refers to the format of spatial arrangement of geometric shapes, which can be composed in the forms of line, cross, triangle, square, and circle. In partition problems, this attribute represents the number of parts the geometric shapes are partitioned into.

\paragraph{Algebra Component and Attributes}

The algebra component serves as the problem \emph{content}; similarly, it is composed of two mathematical attributes. The first attribute indicates the mathematical objects in the problem, including a list of operators and integer constants. The constants range from 1 to 99 and the values of operators are the four elementary operators in arithmetics: ``$+$'', ``$-$'', ``$\times$'', ``$\div$''.

\begin{figure}[t!]
    \centering
    \includegraphics[width=\linewidth]{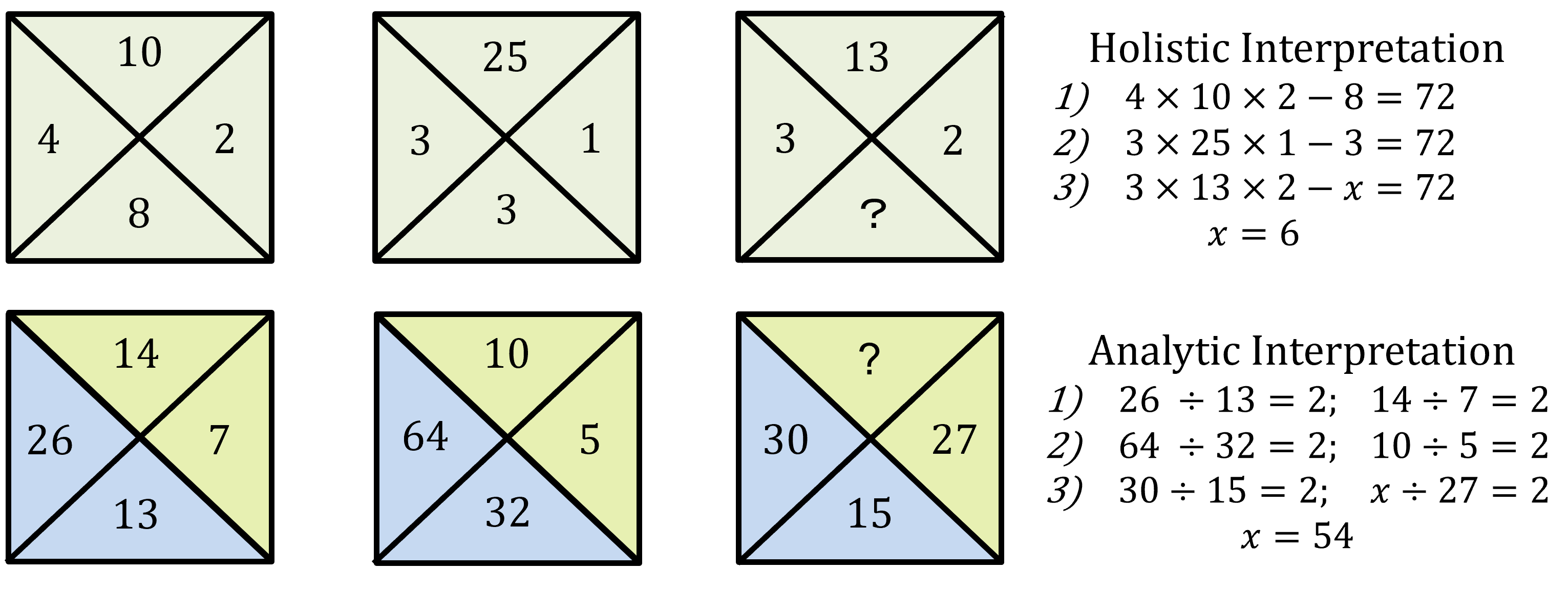}
    \caption{Examples of different algebra components: holistic and analytic interpretation.}
    \label{fig:interpretations}
\end{figure}

The second attribute is the styles of interpretation---holistic view and analytic view, which correspond to two basic thinking styles of human cognition \cite{nisbett2001culture}; see Figure~\ref{fig:interpretations}. Holistic cognition emphasizes attending to the entire information input, while analytic cognition focuses on grouping the input into different sub-parts. From a holistic perspective, all the numbers in a panel are involved in the same calculation process together as a whole. From an analytic perspective, the numbers are grouped as several parts, and each part undergoes an individual calculation process. If the interpretation style is analytic, the numbers in a panel can be divided into 2, 3, or 4 parts. The form of grouping is designed on the basis of human perceptual organization laws, especially the law of similarity and the law of proximity \cite{wertheimer1923laws}: numbers at neighboring or symmetrical positions tend to be organized as a group.

\paragraph{Sampling Math Objects}

Once the layout and the interpretation style are determined, the final step is to sample operators and constants to automatically generate a test.

The space of possible operators given the current problem type, component, and attribute is constrained by the problem context. For holistic problems, this space is subject to the number of available integer positions in each \emph{panel}. For analytic problems, the space is subject to the number of available integer positions in each \emph{group}. Parentheses are further randomly inserted to change the operator precedence; this modification dramatically increases the problem space.

The values of integer constants also need to be adjusted to maintain a consistent difficulty among the generated tests. If the center position of the layout has the space to display numbers, we show the integer constant at the center of each panel as a hint for problem-solving. In other situations where the center position is occupied by lines or other shapes, the algorithm needs to reason about what the constant is and how to calculate such a constant. To make a trade-off of difficulty, the values of constants differ in each panel in the former situation, whereas the underlying constants remain the same among different panels in the latter situation.

\paragraph{Instantiation by Calculation Tree}

\begin{figure}[t!]
    \centering
    \includegraphics[width=\linewidth]{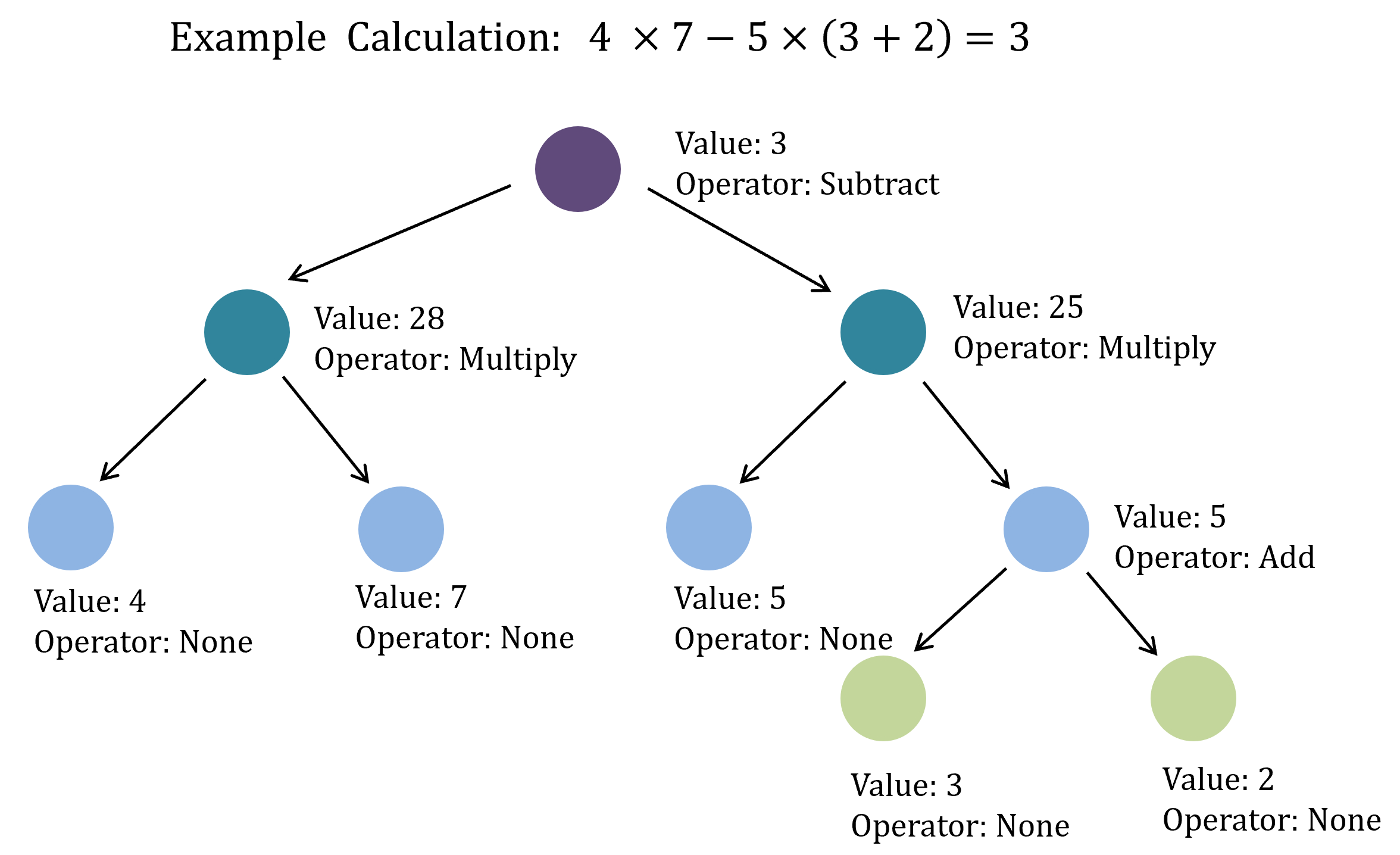}
    \caption{An example of the calculation tree for number generation. The root node is the sampled integer constant.}
    \label{fig:calculation}
\end{figure}

The sampled operators and constant slots are fed into an in-order binary tree to sample numbers for instantiation; see Figure~\ref{fig:calculation} for a detailed example of the generation process. We call this binary tree a ``calculation tree''; the nodes in it have two properties---numeric value and operator. The value of root node is assigned as the already sampled integer constant, while the values of other nodes need to be sampled. The sampling process follows two constraints: (i) the operation between the left-child value and the right-child value under the parent operator will yield the parent value, and (ii) the value is an integer from 1 to 99. The sampling process terminates when all the leaf nodes have qualified values. If a sampling process cannot generate a problem that satisfies all the constraints, it will be terminated and the entire process will be restarted.

\section{Experiment Settings}

We benchmark the proposed \ac{mns} dataset using both predominant neural network models and classic search-based algorithms. Additionally, human performance on the dataset has also been collected.

\subsection{Neural Network Models}

\begin{figure}[t!]
    \centering
    \includegraphics[width=\linewidth]{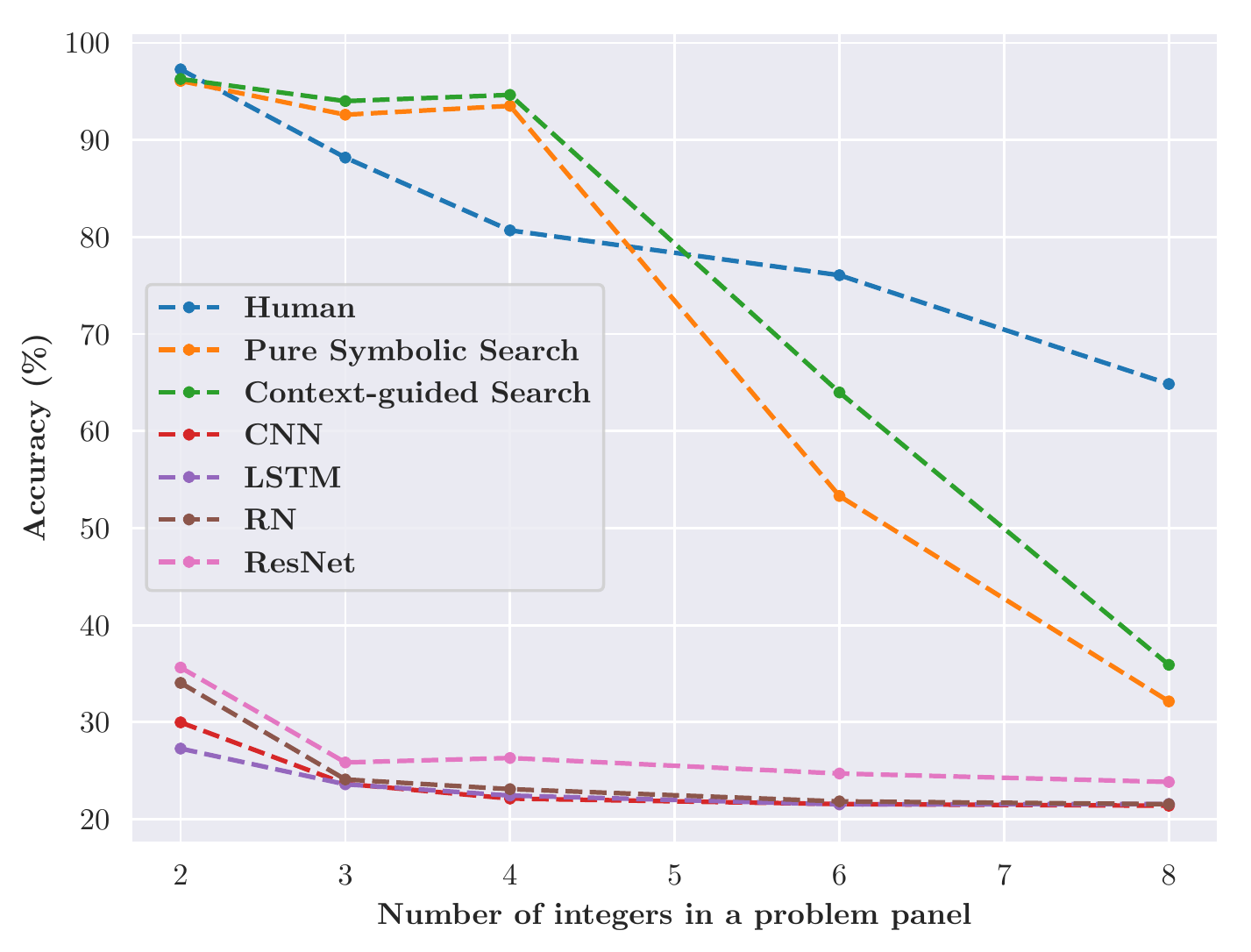}
    \caption{Accuracy w.r.t the number of integers in each panel.}
    \label{fig:number_integer}
\end{figure}

We implement state-of-the-art neural-network-based computer vision models for visual problem-solving \cite{zhang2019raven,barrett2018measuring} and examine their competence on the dataset. Specifically, we compare 4 different baselines: (i) a front-end CNN as feature extractor (CNN), (ii) a popular sequential learning model with a CNN backbone combined with an MLP head (LSTM), (iii) an image classifier based on ResNet \cite{he2016deep}, and (iv) a relational network (RN) \cite{santoro2017simple}. We treat the problem as a classification problem and train all models using the cross-entropy loss. All models are optimized using ADAM \cite{kingma2014adam} and implemented by PyTorch \cite{paszke2017automatic}; see performance in Table~\ref{tab:benchmark} and Figure~\ref{fig:number_integer}.

\paragraph{CNN} In this model, we treat the three panels as a whole and stack them along the channel dimension. Features of the stacked panels are extracted by a CNN model, from which a final answer is predicted.

\paragraph{LSTM} The sequential nature of calculation and the analogical relations among different panels motivate us to choose the representative LSTM model for sequential learning. Similar to ConvLSTM \cite{xingjian2015convolutional}, we feed each panel independently through a small CNN feature extractor and connect them to the input layer of an LSTM network. Image features are iteratively updated in three steps and finally passed to a multi-layer perceptron for prediction.

\paragraph{ResNet} Due to the superior performance in image classification, we also choose to benchmark the dataset using ResNet. The feature extractor used in CNN is now replaced with a ResNet-18 backbone \cite{he2016deep}. We use the publicly available implementation and train the model from random initialization.

\paragraph{RN} Relational network has demonstrated good performance on tasks demanding relational reasoning \cite{santoro2017simple,barrett2018measuring}. Hence, it is natural to examine whether such a relational structure could be beneficial for number sense. We adopt the relational model and feed image features extracted by a CNN. A multi-layer perceptron is used to predict answers based on the relational representation.

\begin{table*}[t!]
    \centering
    \begin{tabular}{ccccccccc}
        \hline
        \multirow{2}{*}{Method} & \multirow{2}{*}{Mean} & \multicolumn{2}{c}{Combination} & \multicolumn{2}{c}{Composition} & \multicolumn{2}{c}{Partition} \\ 
        \cline{3-8} 
                               &                       & Holistic & Analytic               & Holistic & Analytic               & Holistic & Analytic                \\
        \hline\hline
        Pure Symbolic Search    & 52.15\%               & 62.98\%  & 56.83\%                & 22.17\%  & 53.73\%                & 51.29\%  & 71.60\%                 \\
        Context-guided Search  & 56.70\%               & 64.38\%  & 56.08\%                & 29.81\%  & 61.84\%                & 59.70\%  & 67.59\%                 \\
        \hline
        CNN                    & 22.71\%               & 25.25\%  & 19.65\%                & 22.53\%  & 20.07\%                & 24.44\%  & 23.25\%                 \\
        LSTM                   & 22.16\%               & 24.57\%  & 21.10\%                & 22.21\%  & 20.12\%                & 23.36\%  & 23.83\%                 \\
        RN                     & 22.96\%               & 27.05\%  & 20.47\%                & 22.93\%  & 20.27\%                & 25.81\%  & 23.64\%                 \\
        ResNet                 & 25.29\%               & 27.90\%  & 24.22\%                & 23.42\%  & 23.73\%                & 26.61\%  & 27.78\%                 \\
        \hline
        Human                  & 77.58\%               & 66.82\%  & 93.64\%                & 61.36\%  & 78.18\%                & 77.27\%  & 88.18\%
                       \\
        \hline
    \end{tabular}%
    \caption{Performance (accuracy) of different models on the machine number sense dataset.}
    \label{tab:benchmark}
\end{table*}

\subsection{Symbolic Search-Based Models}

We further examine whether the tests can be solved by searching through the problem space. The problem space is fairly large, spanned by various operators, constants, interpretations, shapes, and relations, posing challenges for symbolic search-based models.

We implemented two types of the symbolic search-based models: (i) pure symbolic search, wherein the input is the numbers in each panel, and (ii) context-guided search, taking both the numbers and semantic context information as input. Both the pure symbolic search and context-guided search share similar problem-solving mechanisms: search through the entire problem space until the problem is solved.

Context-guided search only differs from pure symbolic search in two aspects: (i) additional context information may provide heuristics for solving the problem, and (ii) the relative spatial positions of numbers can be inferred from context information, enabling the model to find the correct order of numbers in calculation more quickly. The performance using these two models are shown in Table~\ref{tab:benchmark} and Figure~\ref{fig:number_integer}.

\section{Performance Analysis and Comparison}

\subsection{Analysis of Neural Network Models}

Table~\ref{tab:benchmark} shows how models perform on the \ac{mns} dataset. As shown in Table~\ref{tab:benchmark}, neural networks, unlike search algorithms, perform similarly on different interpretations across all problem types. This observation indicates that by purely learning from the paired image and answer, neural network models are not capable of acquiring the essential cognitive process of perception organization for analytic interpretation. Among all the tested models, ResNet achieves the best performance compared to other neural network models. One possible contribution to the better performance of ResNet may come from its considerable depth, which enables the model to extract more distinct features from the problem images \cite{he2016deep}, helping to discriminate a certain number symbol from others. Although discriminative features on symbols alone may be inadequate for a comprehensive symbolic understanding, it indicates that a strong classifier does help to improve the overall performance.

Figure~\ref{fig:number_integer} shows how model performance changes as the number of integers involved increases. One counter-intuitive observation for neural network models is that the accuracy of problem-solving does not significantly decrease as the number of integers increases. Although the accuracy is the highest in $2$-integer situation for all models, the performance in cases with more integers remain similar. This observation suggests that neural network models share a common processing mechanism that is invariant to the number of integers, qualitatively different from search algorithms.

\subsection{Analysis of Search-based Models}

\begin{figure}[t!]
    \centering
    \includegraphics[width=\linewidth]{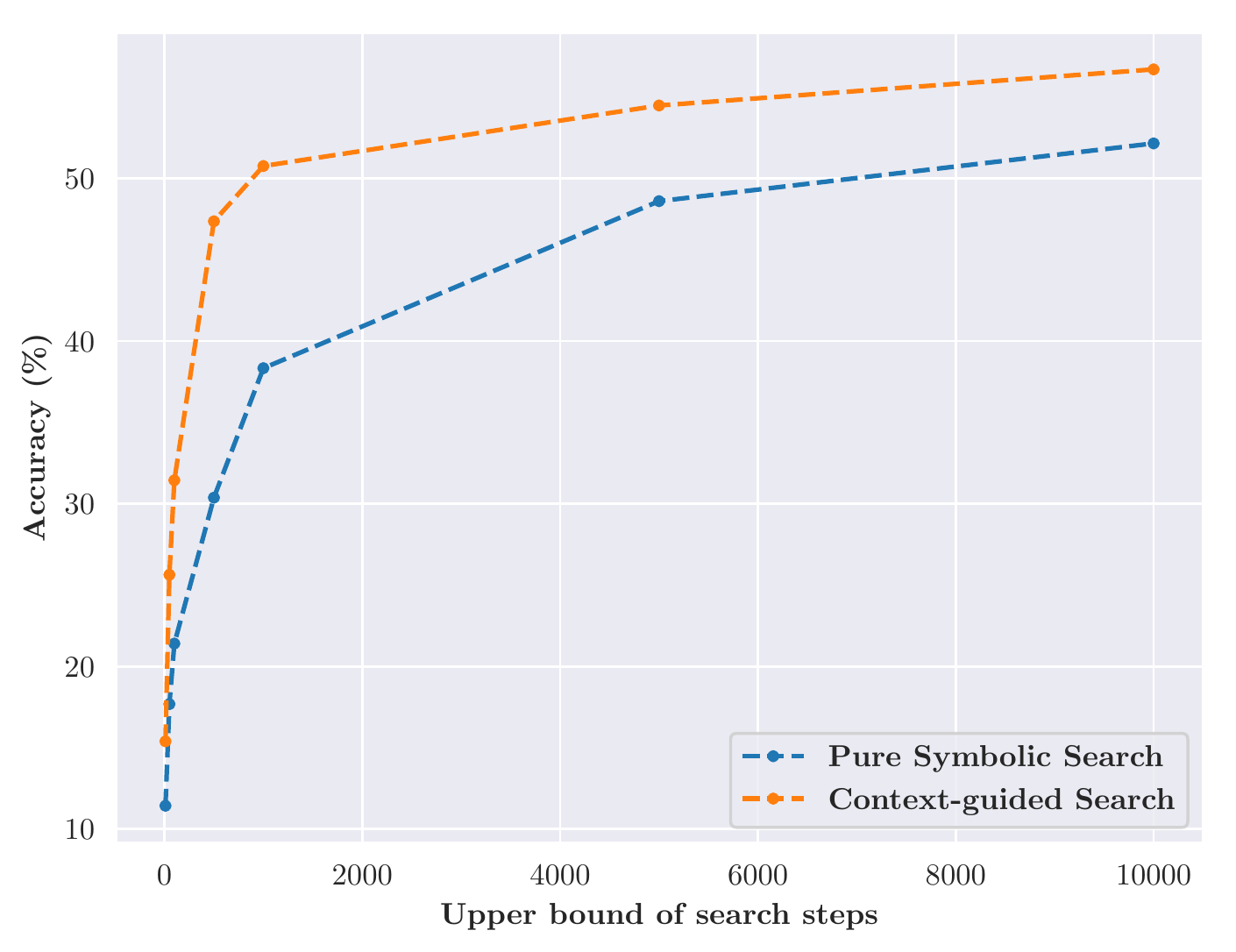}
    \caption{Search performance improves as the number of maximum searching steps increases.}
    \label{fig:searching_step}
\end{figure}

Figure~\ref{fig:searching_step} shows that the accuracy of search algorithms improves as the number of search steps increases, in accordance with the intuition that more trials during problem-solving will lead to a higher chance of success. We observe from Table~\ref{tab:benchmark} that the performance of search algorithms differs between the two styles of interpretations. In combination problems, the algorithms perform better in holistic interpretation. Conversely, in partition and composition problems, the algorithms perform better in analytic interpretation. This observation follows the design of problem layouts: as there are usually more integers in partition and composition problems, it is more expensive to conduct holistic calculations than grouping the integers into several parts for computation. We also note that four numbers could be a turning point for search-based algorithms as the performance drops significantly when there are more than four integers.

Although pure symbolic search is able to solve some problems, context-guided search has, in general, better performance, especially on problems with higher complexity, \eg, $4$-, $6$- and $8$-integer (see Figure~\ref{fig:number_integer}). This difference shows the importance of context information in formulating a suitable organization and representation of problem, avoiding invalid trials of low-possibility circumstances, and finding solutions for complicated problems.

\subsection{Compare Search \emph{vs.} Neural Network}

There are two major differences in performance of search algorithms and neural network models: 
\begin{itemize}[leftmargin=*,noitemsep,nolistsep]
    \item The overall accuracy of neural network models is close to that of pure symbolic search within 100 steps and context-guided search within 50 steps, both of which are relatively small compared to the large problem space. 
    \item The performance of search algorithms varies across different types of problem, different styles of interpretation, and different numbers of integers, in strong contrast to the performance consistency of neural network models. 
\end{itemize}

The underlying reasons for the differences lies in three aspects. First, the representations of number symbols and geometric contexts differ. For search algorithms, the input number symbols are represented as abstract concepts, with clear quantitative meaning and known operational rules, which can be directly fit into each calculation process. Similarly, the context information is given as a high-level semantic concept. In contrast, for neural network models, the input number symbols and geometric contexts are in the form of pixels, so that the models represent the information as a set of extracted features rather than a set of symbolized concepts.

Second, search-based models treat number symbols as independent concepts and process them in a sequential manner, resulting in increased time complexity as the number of integers grows. In contrast, neural network models process visual features in parallel, so that the model performances are invariant to the number of integers.

Third, since the number symbols and geometric context information are fed into search algorithms separately, the ability of search algorithms to separate problem content from problem context is also advantageous than that of neural network models. We argue that being able to separate contents and contexts based on the pixel input is crucial to achieve high performance for neural network models: geometric figures will lose its meaning if deprived of problem contents, and the interpretation of problem contents will also be harder without problem contexts.

\subsection{Compare Human \emph{vs.} Machine Performance}

Compared to computational models, human achieves a significantly higher accuracy in \emph{all} types of problems without extensive training. In our experiments, participants have displayed a superb proficiency in comprehending number symbols and an advanced capacity to learn about the operational relations among numbers from just \emph{two} problem panels.

Unlike neural network models and search algorithms, participants consistently perform better in analytic calculation than in holistic calculation. Similar to search algorithms, the accuracy of participants drops as the amount of number symbols in a problem panel increases. A surprising result is that participants only perform better than search algorithms when there are more than four integers in each problem panel; this counter-intuitive result may be due to the fact that the search-based algorithms can almost search the entire problem space within the step limit when the number of integers is small.

\section{Related Work}

Investigating machine number sense is an important direction that would shed light on many other research topics in the area of artificial intelligence and cognitive science, such as relational reasoning \cite{waltz1999system,santoro2017simple,zhang2019raven}, visual analogy \cite{stafford2001visual,davies2001visual,hill2019learning,zhang2019raven,zhang2019learning}, and concept learning \cite{tenenbaum1999bayesian,fisher2014concept,lake2015human}. Below, we briefly review related work in number sense.

\subsection{Educational Psychology}

To examine the number sense in students' math learning, researchers in the field of educational psychology have been developing and standardizing the diagnostic measurement and intervention training of number sense. A series of tests has been devised to examine different aspects of number sense, which can be classified into two categories---conceptual understanding and procedural operation. For example, the Quantity Discrimination task measures the understanding of the quantitative meaning of number symbols \cite{chard2005using}, and the Number Knowledge Test assesses operational knowledge of numbers (\eg, basic additions and subtractions) with a hierarchy of difficulty \cite{okamoto1996ii}.

\subsection{Artificial Intelligence}

The cognitive ability of machine number sense has not been thoroughly investigated in the field of artificial intelligence. Although mathematical problem-solving has been a research topic with an increasing interest, the focus of previous research work is either on abstract language understanding \cite{kushman2014learning,upadhyay2016annotating,huang2016well,wang2017deep,ling2017program} or general mathematical problem-solving \cite{saxton2019analysing}, leaving out the specific topic of ``machine number sense''. Crucially, almost all the prior work presented mathematical problems in the form of text, which will lead to a sequential (thus simplified) problem-solving process. This process is different from the flexible nature of human cognition shown in mathematics.

\subsection{Machine IQ and Analogy}

Another highly related stream of research focuses on relational and analogical reasoning~\cite{barrett2018measuring,zhang2019raven,zhang2019learning}. Recently, researchers proposed to use deep neural networks to solve Raven's Progressive Matrices (RPM)~\cite{raven1936mental,raven1998raven}. Unlike the proposed number sense challenge, RPM involves a wider range of object relations, such as figure addition, subtraction, and distribution. However, the requirement of number sense in RPM is less demanding than the proposed dataset: only a limited number of objects are involved in each RPM instance and there is no need for decomposition based on the problem context. A similar setting is studied in~\cite{edmonds2020theory,edmonds2019decomposing}, where an agent needs to reason about the open-lock mechanism by generalizing from mechanistically similar but visually different environments. Our work echoes their conclusion that current methods in training deep neural networks do not help the models acquire a generalizable representation.

To solve RPM problems by computational modeling, previous works also adopted knowledge-based methods, outlining analogical reasoning in a predefined manner. For example, some researchers established structural mappings between RPM image panels, which directly processed the visual objects as a set of abstract concepts and variation rules \cite{Lovett2010Solving,lovett2010structure,lovett2017modeling}. With the provided symbolic concepts and relation rules, the RPM problems can be converted to search problems, leading to an optimal problem-solving accuracy outperforming human. Inspired by these previous investigations, the search-based algorithms in our work were also equipped with prior knowledge of numbers and calculations, as well as analogical mappings between image panels. However, as the arithmetic problems in the \ac{mns} dataset possess much larger search space than RPM problems, it is hard and consumptive to solve these problems simply by a pure symbolic search regardless of contextual information. An additional perception module could accelerate the search process with some heuristics from problem contexts, addressing the challenge to some degree; but an apparent gap between search efficiency and human performance still exists. Compared with the previous work, our work reveals the insufficiency of knowledge-based methods in integrating provided knowledge, problem content, and contextual information to conduct human-like adaptive problem-solving. 

\section{Discussions and Conclusion}

In this paper, we propose a dataset generated by \acf{aog} to examine the \emph{machine number sense}. Specifically, we evaluate machines' understanding of abstract number symbols and competence of context-based problem-solving. Compared to simple symbolic search-based models, the poor performance of neural network models suggests its insufficiency in symbolic processing and concept understanding, as well as its difficulty in combining content and context to solve problems flexibly. 

The dataset and experiments have left room for improvements and brought up inspirations for future work. The critical challenges are how to \emph{emerge} symbolic concepts directly from pixels using minimal supervisions, how to extract \emph{meaningful} relations from the contextual information, and how to reason and make inductions based on these concepts and relations. As the experiments indicated, fusing neural network models' strong capacity of visual feature extraction in large-scale data processing and search-based algorithms' explicit knowledge structure in fit-for-purpose problem-solving may be an effective method for relational and abstract reasoning; the integration of data-driven and knowledge-based methods will complement each other.

\subsection*{Acknowledgement}

The authors thank Prof. Hongjing Lu at UCLA Psychology Department for helpful discussions. This work reported herein is supported by MURI ONR N00014-16-1-2007, DARPA XAI N66001-17-2-4029, ONR N00014-19-1-2153, and an NVIDIA GPU donation grant.

\bibliographystyle{aaai}
\bibliography{aaai20}

\begin{thebibliography}{}

\bibitem[\protect\citeauthoryear{Barrett \bgroup et al\mbox.\egroup
  }{2018}]{barrett2018measuring}
Barrett, D.~G.; Hill, F.; Santoro, A.; Morcos, A.~S.; and Lillicrap, T.
\newblock 2018.
\newblock Measuring abstract reasoning in neural networks.
\newblock {\em arXiv preprint arXiv:1807.04225}.

\bibitem[\protect\citeauthoryear{Bobis}{1996}]{bobis1996visualisation}
Bobis, J.
\newblock 1996.
\newblock Visualisation and the development of number sense with kindergarten
  children.
\newblock {\em Children’s number learning: A research monograph of MERGA/AAMT
  Adelaide: Australian Association of Mathematics teachers}.

\bibitem[\protect\citeauthoryear{Brannon and
  Terrace}{1998}]{brannon1998ordering}
Brannon, E.~M., and Terrace, H.~S.
\newblock 1998.
\newblock Ordering of the numerosities 1 to 9 by monkeys.
\newblock {\em Science} 282(5389):746--749.

\bibitem[\protect\citeauthoryear{Cantlon and Brannon}{2007}]{cantlon2007much}
Cantlon, J.~F., and Brannon, E.~M.
\newblock 2007.
\newblock How much does number matter to a monkey (macaca mulatta)?
\newblock {\em Journal of Experimental Psychology: Animal Behavior Processes}
  33(1):32.

\bibitem[\protect\citeauthoryear{Cattell}{1963}]{cattell1963theory}
Cattell, R.~B.
\newblock 1963.
\newblock Theory of fluid and crystallized intelligence: A critical experiment.
\newblock {\em Journal of educational psychology} 54(1):1.

\bibitem[\protect\citeauthoryear{Chard \bgroup et al\mbox.\egroup
  }{2005}]{chard2005using}
Chard, D.~J.; Clarke, B.; Baker, S.; Otterstedt, J.; Braun, D.; and Katz, R.
\newblock 2005.
\newblock Using measures of number sense to screen for difficulties in
  mathematics: Preliminary findings.
\newblock {\em Assessment for Effective Intervention} 30(2):3--14.

\bibitem[\protect\citeauthoryear{Cobb \bgroup et al\mbox.\egroup
  }{1991}]{cobb1991assessment}
Cobb, P.; Wood, T.; Yackel, E.; Nicholls, J.; Wheatley, G.; Trigatti, B.; and
  Perlwitz, M.
\newblock 1991.
\newblock Assessment of a problem-centered second-grade mathematics project.
\newblock {\em Journal for research in mathematics education}  3--29.

\bibitem[\protect\citeauthoryear{Davies and Goel}{2001}]{davies2001visual}
Davies, J., and Goel, A.~K.
\newblock 2001.
\newblock Visual analogy in problem solving.
\newblock In {\em IJCAI}.

\bibitem[\protect\citeauthoryear{Davis and
  P{\'e}russe}{1988}]{davis1988numerical}
Davis, H., and P{\'e}russe, R.
\newblock 1988.
\newblock Numerical competence in animals: Definitional issues, current
  evidence, and a new research agenda.
\newblock {\em Behavioral and Brain Sciences} 11(4):561--579.

\bibitem[\protect\citeauthoryear{Dehaene, Dehaene-Lambertz, and
  Cohen}{1998}]{dehaene1998abstract}
Dehaene, S.; Dehaene-Lambertz, G.; and Cohen, L.
\newblock 1998.
\newblock Abstract representations of numbers in the animal and human brain.
\newblock {\em Trends in neurosciences} 21(8):355--361.

\bibitem[\protect\citeauthoryear{Dehaene}{2011}]{dehaene2011number}
Dehaene, S.
\newblock 2011.
\newblock {\em The number sense: How the mind creates mathematics}.
\newblock OUP USA.

\bibitem[\protect\citeauthoryear{Edmonds \bgroup et al\mbox.\egroup
  }{2019}]{edmonds2019decomposing}
Edmonds, M.; Qi, S.; Zhu, Y.; Kubricht, J.; Zhu, S.-C.; and Lu, H.
\newblock 2019.
\newblock Decomposing human causal learning: Bottom-up associative learning and
  top-down schema reasoning.
\newblock In {\em CogSci}.

\bibitem[\protect\citeauthoryear{Edmonds \bgroup et al\mbox.\egroup
  }{2020}]{edmonds2020theory}
Edmonds, M.; Ma, X.; Qi, S.; Zhu, Y.; Lu, H.; and Zhu, S.-C.
\newblock 2020.
\newblock Theory-based causal transfer: Integrating instance-level induction
  and abstract-level structure learning.
\newblock In {\em AAAI}.

\bibitem[\protect\citeauthoryear{Fisher, Pazzani, and
  Langley}{2014}]{fisher2014concept}
Fisher, D.~H.; Pazzani, M.~J.; and Langley, P.
\newblock 2014.
\newblock {\em Concept formation: Knowledge and experience in unsupervised
  learning}.
\newblock Morgan Kaufmann.

\bibitem[\protect\citeauthoryear{Gallistel}{1989}]{gallistel1989animal}
Gallistel, C.~R.
\newblock 1989.
\newblock Animal cognition: The representation of space, time and number.
\newblock {\em Annual review of psychology} 40(1):155--189.

\bibitem[\protect\citeauthoryear{Gallistel}{1990}]{gallistel1990organization}
Gallistel, C.~R.
\newblock 1990.
\newblock {\em The organization of learning.}
\newblock The MIT Press.

\bibitem[\protect\citeauthoryear{Gallistel}{2003}]{Gallistel2003Animal}
Gallistel, C.~R.
\newblock 2003.
\newblock Animal cognition: the representation of space, time and number.
\newblock {\em Annual Review of Psychology} 40(40):155--189.

\bibitem[\protect\citeauthoryear{He \bgroup et al\mbox.\egroup
  }{2016}]{he2016deep}
He, K.; Zhang, X.; Ren, S.; and Sun, J.
\newblock 2016.
\newblock Deep residual learning for image recognition.
\newblock In {\em CVPR}.

\bibitem[\protect\citeauthoryear{Heinze, Star, and
  Verschaffel}{2009}]{heinze2009flexible}
Heinze, A.; Star, J.~R.; and Verschaffel, L.
\newblock 2009.
\newblock Flexible and adaptive use of strategies and representations in
  mathematics education.
\newblock {\em ZDM} 41(5):535--540.

\bibitem[\protect\citeauthoryear{Hill \bgroup et al\mbox.\egroup
  }{2019}]{hill2019learning}
Hill, F.; Santoro, A.; Barrett, D. G.~T.; Morcos, A.~S.; and Lillicrap, T.~P.
\newblock 2019.
\newblock Learning to make analogies by contrasting abstract relational
  structure.
\newblock {\em ArXiv} abs/1902.00120.

\bibitem[\protect\citeauthoryear{Huang \bgroup et al\mbox.\egroup
  }{2016}]{huang2016well}
Huang, D.; Shi, S.; Lin, C.-Y.; Yin, J.; and Ma, W.-Y.
\newblock 2016.
\newblock How well do computers solve math word problems? large-scale dataset
  construction and evaluation.
\newblock In {\em ACL}.

\bibitem[\protect\citeauthoryear{Jacob and Nieder}{2008}]{jacob2008abc}
Jacob, S.~N., and Nieder, A.
\newblock 2008.
\newblock The abc of cardinal and ordinal number representations.
\newblock {\em Trends in cognitive sciences} 12(2):41--43.

\bibitem[\protect\citeauthoryear{Kalchman, Moss, and
  Case}{2001}]{kalchman2001psychological}
Kalchman, M.; Moss, J.; and Case, R.
\newblock 2001.
\newblock Psychological models for the development of mathematical
  understanding: Rational numbers and functions.
\newblock {\em Cognition and instruction: Twenty-five years of progress}
  1--38.

\bibitem[\protect\citeauthoryear{Kingma and Ba}{2014}]{kingma2014adam}
Kingma, D.~P., and Ba, J.
\newblock 2014.
\newblock Adam: A method for stochastic optimization.
\newblock {\em arXiv preprint arXiv:1412.6980}.

\bibitem[\protect\citeauthoryear{Knoblich \bgroup et al\mbox.\egroup
  }{1999}]{knoblich1999constraint}
Knoblich, G.; Ohlsson, S.; Haider, H.; and Rhenius, D.
\newblock 1999.
\newblock Constraint relaxation and chunk decomposition in insight problem
  solving.
\newblock {\em Journal of Experimental Psychology: Learning, memory, and
  cognition} 25(6):1534.

\bibitem[\protect\citeauthoryear{Kushman \bgroup et al\mbox.\egroup
  }{2014}]{kushman2014learning}
Kushman, N.; Artzi, Y.; Zettlemoyer, L.; and Barzilay, R.
\newblock 2014.
\newblock Learning to automatically solve algebra word problems.
\newblock In {\em ACL}.

\bibitem[\protect\citeauthoryear{Lake, Salakhutdinov, and
  Tenenbaum}{2015}]{lake2015human}
Lake, B.~M.; Salakhutdinov, R.; and Tenenbaum, J.~B.
\newblock 2015.
\newblock Human-level concept learning through probabilistic program induction.
\newblock {\em Science} 350(6266):1332--1338.

\bibitem[\protect\citeauthoryear{Ling \bgroup et al\mbox.\egroup
  }{2017}]{ling2017program}
Ling, W.; Yogatama, D.; Dyer, C.; and Blunsom, P.
\newblock 2017.
\newblock Program induction by rationale generation: Learning to solve and
  explain algebraic word problems.
\newblock {\em arXiv preprint arXiv:1705.04146}.

\bibitem[\protect\citeauthoryear{Lovett and Forbus}{2017}]{lovett2017modeling}
Lovett, A., and Forbus, K.
\newblock 2017.
\newblock Modeling visual problem solving as analogical reasoning.
\newblock {\em Psychological review} 124(1):60.

\bibitem[\protect\citeauthoryear{Lovett \bgroup et al\mbox.\egroup
  }{2010}]{Lovett2010Solving}
Lovett, A.; Tomai, E.; Forbus, K.; and Usher, J.
\newblock 2010.
\newblock Solving geometric analogy problems through two-stage analogical
  mapping.
\newblock {\em Cognitive Science} 33(7):1192--1231.

\bibitem[\protect\citeauthoryear{Lovett, Forbus, and
  Usher}{2010}]{lovett2010structure}
Lovett, A.; Forbus, K.; and Usher, J.
\newblock 2010.
\newblock A structure-mapping model of raven's progressive matrices.
\newblock In {\em CogSci}.

\bibitem[\protect\citeauthoryear{Nieder and
  Dehaene}{2009}]{nieder2009representation}
Nieder, A., and Dehaene, S.
\newblock 2009.
\newblock Representation of number in the brain.
\newblock {\em Annual review of neuroscience} 32:185--208.

\bibitem[\protect\citeauthoryear{Nisbett \bgroup et al\mbox.\egroup
  }{2001}]{nisbett2001culture}
Nisbett, R.~E.; Peng, K.; Choi, I.; and Norenzayan, A.
\newblock 2001.
\newblock Culture and systems of thought: holistic versus analytic cognition.
\newblock {\em Psychological review} 108(2):291.

\bibitem[\protect\citeauthoryear{Okamoto and Case}{1996}]{okamoto1996ii}
Okamoto, Y., and Case, R.
\newblock 1996.
\newblock Ii. exploring the microstructure of children's central conceptual
  structures in the domain of number.
\newblock {\em Monographs of the Society for research in Child Development}
  61(1-2):27--58.

\bibitem[\protect\citeauthoryear{Paszke \bgroup et al\mbox.\egroup
  }{2017}]{paszke2017automatic}
Paszke, A.; Gross, S.; Chintala, S.; Chanan, G.; Yang, E.; DeVito, Z.; Lin, Z.;
  Desmaison, A.; Antiga, L.; and Lerer, A.
\newblock 2017.
\newblock Automatic differentiation in pytorch.
\newblock In {\em ICLR}.

\bibitem[\protect\citeauthoryear{Raven and Court}{1998}]{raven1998raven}
Raven, J.~C., and Court, J.~H.
\newblock 1998.
\newblock {\em Raven's progressive matrices and vocabulary scales}.
\newblock Oxford pyschologists Press.

\bibitem[\protect\citeauthoryear{Raven}{1936}]{raven1936mental}
Raven, J.~C.
\newblock 1936.
\newblock Mental tests used in genetic studies: The performance of related
  individuals on tests mainly educative and mainly reproductive.
\newblock Master's thesis, University of London.

\bibitem[\protect\citeauthoryear{Santoro \bgroup et al\mbox.\egroup
  }{2017}]{santoro2017simple}
Santoro, A.; Raposo, D.; Barrett, D.~G.; Malinowski, M.; Pascanu, R.;
  Battaglia, P.; and Lillicrap, T.
\newblock 2017.
\newblock A simple neural network module for relational reasoning.
\newblock In {\em NeurIPS}.

\bibitem[\protect\citeauthoryear{Saxton \bgroup et al\mbox.\egroup
  }{2019}]{saxton2019analysing}
Saxton, D.; Grefenstette, E.; Hill, F.; and Kohli, P.
\newblock 2019.
\newblock Analysing mathematical reasoning abilities of neural models.
\newblock {\em arXiv preprint arXiv:1904.01557}.

\bibitem[\protect\citeauthoryear{Schneider \bgroup et al\mbox.\egroup
  }{2017}]{schneider2017associations}
Schneider, M.; Beeres, K.; Coban, L.; Merz, S.; Susan~Schmidt, S.; Stricker,
  J.; and De~Smedt, B.
\newblock 2017.
\newblock Associations of non-symbolic and symbolic numerical magnitude
  processing with mathematical competence: A meta-analysis.
\newblock {\em Developmental science} 20(3):e12372.

\bibitem[\protect\citeauthoryear{Simpson and
  Weiner}{1989}]{dictionary1989oxford}
Simpson, J., and Weiner, E.
\newblock 1989.
\newblock Oxford english dictionary.
\newblock {\em Dictionary, Oxford English}.

\bibitem[\protect\citeauthoryear{Stafford}{2001}]{stafford2001visual}
Stafford, B.~M.
\newblock 2001.
\newblock {\em Visual analogy: Consciousness as the art of connecting}.
\newblock MIT press.

\bibitem[\protect\citeauthoryear{Starkey and
  Cooper}{1980}]{starkey1980perception}
Starkey, P., and Cooper, R.~G.
\newblock 1980.
\newblock Perception of numbers by human infants.
\newblock {\em Science} 210(4473):1033--1035.

\bibitem[\protect\citeauthoryear{Taylor}{1818}]{taylor1818nicomachean}
Taylor, T.
\newblock 1818.
\newblock {\em The Nicomachean ethics}.
\newblock AJ Valpy.

\bibitem[\protect\citeauthoryear{Temple and Posner}{1998}]{temple1998brain}
Temple, E., and Posner, M.~I.
\newblock 1998.
\newblock Brain mechanisms of quantity are similar in 5-year-old children and
  adults.
\newblock {\em PNAS} 95(13):7836--7841.

\bibitem[\protect\citeauthoryear{Tenenbaum}{1999}]{tenenbaum1999bayesian}
Tenenbaum, J.~B.
\newblock 1999.
\newblock Bayesian modeling of human concept learning.
\newblock In {\em NeurIPS}.

\bibitem[\protect\citeauthoryear{Upadhyay and
  Chang}{2016}]{upadhyay2016annotating}
Upadhyay, S., and Chang, M.-W.
\newblock 2016.
\newblock Annotating derivations: A new evaluation strategy and dataset for
  algebra word problems.
\newblock {\em arXiv preprint arXiv:1609.07197}.

\bibitem[\protect\citeauthoryear{Waltz \bgroup et al\mbox.\egroup
  }{1999}]{waltz1999system}
Waltz, J.~A.; Knowlton, B.~J.; Holyoak, K.~J.; Boone, K.~B.; Mishkin, F.~S.;
  de~Menezes~Santos, M.; Thomas, C.~R.; and Miller, B.~L.
\newblock 1999.
\newblock A system for relational reasoning in human prefrontal cortex.
\newblock {\em Psychological science} 10(2):119--125.

\bibitem[\protect\citeauthoryear{Wang, Liu, and Shi}{2017}]{wang2017deep}
Wang, Y.; Liu, X.; and Shi, S.
\newblock 2017.
\newblock Deep neural solver for math word problems.
\newblock In {\em EMNLP}.

\bibitem[\protect\citeauthoryear{Wertheimer}{1923}]{wertheimer1923laws}
Wertheimer, M.
\newblock 1923.
\newblock Laws of organization in perceptual forms.
\newblock {\em A source book of Gestalt Psychology}.

\bibitem[\protect\citeauthoryear{Wiese}{2003}]{wiese2003numbers}
Wiese, H.
\newblock 2003.
\newblock {\em Numbers, language, and the human mind}.
\newblock Cambridge University Press.

\bibitem[\protect\citeauthoryear{Wynn}{1992}]{wynn1992addition}
Wynn, K.
\newblock 1992.
\newblock Addition and subtraction by human infants.
\newblock {\em Nature} 358(6389):749.

\bibitem[\protect\citeauthoryear{Xingjian \bgroup et al\mbox.\egroup
  }{2015}]{xingjian2015convolutional}
Xingjian, S.; Chen, Z.; Wang, H.; Yeung, D.-Y.; Wong, W.-K.; and Woo, W.-c.
\newblock 2015.
\newblock Convolutional lstm network: A machine learning approach for
  precipitation nowcasting.
\newblock In {\em NeurIPS}.

\bibitem[\protect\citeauthoryear{Zhang \bgroup et al\mbox.\egroup
  }{2019a}]{zhang2019raven}
Zhang, C.; Gao, F.; Jia, B.; Zhu, Y.; and Zhu, S.-C.
\newblock 2019a.
\newblock Raven: A dataset for relational and analogical visual reasoning.
\newblock In {\em CVPR}.

\bibitem[\protect\citeauthoryear{Zhang \bgroup et al\mbox.\egroup
  }{2019b}]{zhang2019learning}
Zhang, C.; Jia, B.; Gao, F.; Zhu, Y.; Lu, H.; and Zhu, S.-C.
\newblock 2019b.
\newblock Learning perceptual inference by contrasting.
\newblock In {\em NeurIPS}.

\bibitem[\protect\citeauthoryear{Zhu and Mumford}{2007}]{zhu2007stochastic}
Zhu, S.-C., and Mumford, D.
\newblock 2007.
\newblock A stochastic grammar of images.
\newblock {\em Foundations and Trends in Computer Graphics and Vision}
  2(4):259--362.

\end{thebibliography}

\end{document}